# SDFN: Segmentation-based Deep Fusion Network for Thoracic Disease Classification in Chest X-ray Images

Han Liu, Lei Wang, Yandong Nan, Faguang Jin, Qi Wang, and Jiantao Pu

*Abstract*—This study aims to automatically diagnose thoracic diseases depicted on the chest x-ray (CXR) images using deep convolutional neural networks. The existing methods generally used the entire CXR images for training purposes, but this strategy may suffer from two drawbacks. First, potential misalignment or the existence of irrelevant objects in the entire CXR images may cause unnecessary noise and thus limit the network performance. Second, the relatively low image resolution caused by the resizing operation, which is a common pre-processing procedure for training neural networks, may lead to the loss of image details, making it difficult to detect pathologies with small lesion regions. To address these issues, we present a novel method termed as segmentation-based deep fusion network (SDFN), which leverages the domain knowledge and the higher-resolution information of local lung regions. Specifically, the local lung regions were identified and cropped by the Lung Region Generator (LRG). Two CNN-based classification models were then used as feature extractors to obtain the discriminative features of the entire CXR images and the cropped lung region images. Lastly, the obtained features were fused by the feature fusion module for disease classification. Evaluated by the NIH benchmark split on the Chest X-ray 14 Dataset, our experimental result demonstrated that the developed method achieved more accurate disease classification compared with the available approaches via the receiver operating characteristic (ROC) analyses. It was also found that the SDFN could localize the lesion regions more precisely as compared to the traditional method.

## I. INTRODUCTION

Chest X-ray (CXR) is a very common, non-invasive radiology examination, which has been widely used to screen a variety of thoracic diseases, such as pneumonia, effusion, cancer, and emphysema. In clinical practice, the CXR images are typically interpreted by radiologists and this process is time-consuming and prone to subjective assessment errors [1]. Hence, it is always desirable to have a reliable computer tool that can efficiently and accurately aid in the diagnosis and detection of the diseases depicted on CXR images. As an emerging technology in machine learning, deep learning has demonstrated remarkable strength in a variety tasks of medical image analyses, such as disease classification [2-5], lesion segmentation [6-8], and registration [9-10]. With the recent availability of the Chest X-ray 14 Dataset [11], many deep learning approaches [11-16] have been proposed to automatically diagnose the thoracic diseases in CXR images. Most of the approaches [11-14] used the entire CXR images for training purpose, but this strategy may suffer from two drawbacks. First, the potential artifacts in the entire images, such as image misalignment and irrelevant objects (e.g. medical devices) as shown in Figure 1, may cause unnecessary noise and thus limit the network performance. Second, training deep neural networks typically requires resizing original images to a smaller resolution to allow a higher computational efficiency. However, this resizing process significantly reduces the image resolution and unavoidably leads to the loss of image details, which may be crucial for detecting the pathologies with small lesions. Realizing these drawbacks, some recent studies have proposed to train the networks with higher resolution images: Guan et al. [15] introduced an attention guided convolutional neural network (AG-CNN), where the most discriminative local regions obtained by the class activation maps (CAMs) were considered. However, the CAMs may not correctly indicate the lesion regions and thus lead to unreliable classification. In addition, Gündel et al. [16] proposed to add two extra convolution layers at the beginning of the DenseNet [17] so that the modified network could be trained with full-resolution CXR images. However, limited performance improvement was achieved.

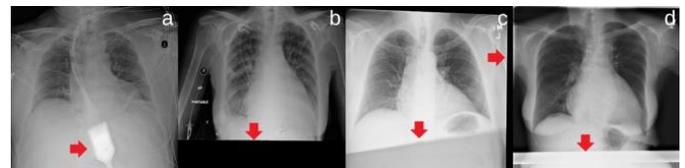

Fig. 1. Examples illustrating the entire CXR images in the Chest X-ray 14 Dataset, where (a) has an irrelevant object and (b)-(d) have misalignment as indicated by the arrows.

In this paper, we present a novel method termed as Segmentation-based Deep Fusion Network (SDFN), which also leverages the higher resolution images: the lung regions. The underlying idea was inspired by the domain knowledge that the thoracic diseases are typically limited within the lung regions. If the lung regions are identified, the networks can be trained using higher resolution images with reduced noisy regions (i.e., misalignment and irrelevant objects). Meanwhile, since the thoracic diseases may also be associated with some other contextual characteristics outside of the lung regions, the entire

The manuscript was submitted on Aug 28st, 2018. This work is supported by National Institutes of Health (NIH) (Grant No. R21CA197493, R01HL096613) and Jiangsu Natural Science Foundation (Grant No. BK20170391).

H. Liu, L. Wang, Qi Wang, and J. Pu are with the Department of Radiology and Bioengineering, University of Pittsburgh, Pittsburgh, PA, 15213. (email: hanliu0621@gmail.com, lew90@pitt.edu, puj@upmc.edu)

Y. Nan, and F. Jin are with the Department of Respiratory and Critical Care Medicine, Tangdu Hospital, Xi'an, China, 710038. (email: 13709205538@163.com, jinfag@fmmu.edu.cn)

CXR images were also considered. By jointly learning the features from the entire CXR and the local lung region, the proposed method consistently improved the classification performance compared to the traditional method. A detailed description of the implementation and the experimental results follows.

## II. MATERIALS AND METHODS

### A. Image datasets

Two publicly available datasets were used in this study. The first dataset is the JSRT Dataset [18], which contains 154 nodule and 93 non-nodule CXR images. Each image has a resolution of 2,048×2,048 pixels with a 12-bit density range. The masks for the lung regions are also publicly available at [19]. The primary objective of JSRT Dataset was to train a lung segmentation model that automatically extract the lung regions from the entire CXR images. The second dataset is the Chest X-ray 14 Dataset, which is comprised of 112,120 frontal-view chest X-ray images acquired on 30,805 unique patients and labeled up to 14 diseases. A total of 984 bounding boxes for 8 pathologies were annotated by a board certified radiologist. The benchmark splits of training and testing sets are publicly available at the NIH website[1]. We used the Chest X-ray 14 Dataset to develop and validate the classification performance of the proposed SDFN. Detailed descriptions of these two datasets can be found in [11,18].

### B. Segmentation-based deep fusion network

The proposed SDFN is a dual-pathway CNN that consists of a lung region generator, two feature extractors respectively for the entire CXR images and local lung region images, and a feature fusion module. The whole pipeline of the SDFN is shown in Figure 2. Given a CXR image, the associated lung region image was obtained by the Lung Region Generator (LRG). Two 121-layer DenseNets were fine-tuned and used as feature extractors to obtain the discriminative features of the entire CXR image and local lung region image. Lastly, the feature fusion module concatenates the global average pooling layers of the two feature extractors and is fine-tuned for the final prediction.

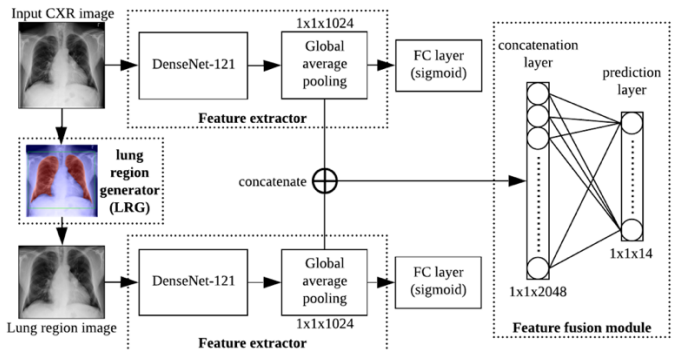

Fig. 2. The pipeline of the proposed SDFN. *FC* represent *fully-connected*.

### B.1. Multi-label setup

Thoracic disease classification is a multi-label classification problem. The input of the proposed SDFN is an entire CXR image and the output is a C-dimension vector $L = [l_1, l_2, l_3, ..., l_C]$, where $C = 14$ and $l_c \in \{0,1\}$ (i.e., '0' and '1' represent absence or presence of a pathology). The pathologies in $L$ are in the order of atelectasis, cardiomegaly, effusion, infiltration, mass, nodule, pneumonia, pneumothorax, consolidation, edema, emphysema, fibrosis, pleural thickening and hernia.

### B.2. Lung region generator (LRG)

The LRG consists of two stages: (1) the lung segmentation model and (2) the post-processing procedures. Given a CXR image, the lung regions were initially predicted by a segmentation model. To further obtain a bounding box that fully enclosed the lung regions, several post-processing procedures were developed based on the anatomical characteristics of lungs.

### B.2.1 Lung segmentation model

The U-net [20] is a well-established convolutional network architecture for fast and precise image segmentation. Recently, a modified U-net [21] trained on the JSRT Dataset has been successfully used to segment the lung regions on the CXR images. To enable automatic lung region extraction for the Chest X-ray 14 Dataset, we followed the same architecture in [21] and retrained this model using the JSRT Dataset. During training, the input CXR images and the lung masks were resized to 256×256 pixels and augmented by affine transformations, such as rotation, shifting, and zooming. We used the Adam optimizer with an initial learning rate of 1e-3 on mini-batch of size 8 and set the maximum number of epochs as 100. The pixel-wise cross-entropy loss function was minimized for optimization and the model with the smallest validation loss was saved.

### B.2.2 Post-processing procedures

As the examples in Figure 3(a) and Figure 4(a) showed, the existence of diseases and/or poor image quality may lead to undesirable segmentation, which may contain false positive and/or false negative regions. To address these issues, we propose to post-process the segmentation results based on the anatomical characteristics of lungs.

For example, the lungs consist of the left and right lungs. Therefore, when the segmentation result contained more than two regions, there were typically false positive identifications, as shown in Figure 3(a). These false positive regions were often caused by the external air or intestine gas, and thus located farther away from the image center compared to the lungs. Hence, we computed the distances between the image center and the centroid of each segmentation region, and only kept the two regions with the shortest distances, as shown in Figure 3(b).

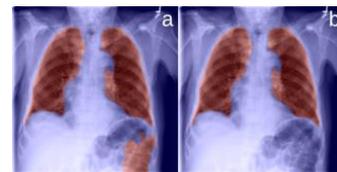

Fig. 3. Remove the false positive segments.

---
[1] https://nihcc.app.box.com/v/ChestXray-NIHCC

For a healthy subject, the left and right lungs typically do not vary significantly in size. Also, due to the CXR scan alignment, the locations of the left and the right lungs are roughly symmetric along the vertical centerline of the image. These characteristics were used to identify the lung regions from the segmentation results with false negative regions, as shown in Figure 4. Specifically, when there were two regions in the segmentation result and their areas were significantly different, we removed the smaller region. The difference was considered to be significant if the smaller region had an area less than one third of the larger region. Thereafter, we mirrored the larger region with respect to the vertical centerline of the image and applied a bounding box for the mirrored region. Lastly, each boundary of the bounding box was expanded with a certain margin (i.e., 15 pixels along top, left and right and 20 pixels along bottom) to assure a fully inclusion of the features around the lung boundaries. In this study, we particularly expanded the box more along the bottom direction in order to include the subtle characteristics located at the bottom of the lungs such as the blunting costophrenic angles. The expanded bounding boxes were cropped from the original CXR images and used as the local lung region images.

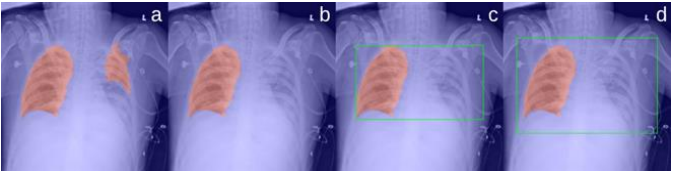

Fig. 4. Illustrating the procedures to create a bounding box for the segmentation results associated with false negative regions. (a) the lung segmentation results (in red), where part of the lungs is not identified, (b) the removal of the smaller regions, (c) the bounding box (in green) of the lung regions after the application of a mirroring operation, and (d) the expansion of the bounding box.

*B.3. Feature extractors*

Due to the relatively higher resolution of the lung region images, the associated features may better describe the thoracic diseases with smaller pathological lesions. Meanwhile, the entire CXR images may include contextual features that are associated with the thoracic diseases but located outside of the lung region. To extract these distinct features, we trained two fine-tuned 121-layer DenseNets on the entire CXR images and the lung region images separately. The DenseNet was used because (1) it has several advantages over other classic networks [25-28] such as alleviating the vanishing-gradient problem, strengthening feature propagation and reuse, and requiring less computation [18], (2) the fine-tuned DenseNet has been reported to be an effective method for thoracic disease classification [13] and thus can be used as a baseline for performance comparison.

In implementation, each DenseNet was fine-tuned by adding a global average pooling layer on top of the last convolution layer, followed by a 14-dimension FC layer. Each output of the FC layer was normalized to [0,1] using a sigmoid activation function, defined by Eq. (1).

$$\tilde{y}_i = \frac{1}{1+e^{-FC_i}} \quad (1)$$

where $\tilde{y}_i$ is the predicted probability of $i_{th}$ class and $FC_i$ is the $i_{th}$ element at the fully-connected layer.

The binary cross-entropy (BCE) loss function, as defined by Eq. (2), was minimized to obtain optimal classification performances.

$$BCE(y, \tilde{y}) = -\frac{1}{N}\sum_{i=1}^{N}[y_i \log(\tilde{y}_i) + (1-y_i)\log(1-\tilde{y}_i)] \quad (2)$$

where $N$ is the number of classes (i.e., $N = 14$), $y_i$ is the ground truth and $\tilde{y}_i$ is the predicted probability.

During training, we initialized the DenseNet-121 with the weights pre-trained on the ImageNet [22]. The input images were resized to 224×224 pixels and augmented by random horizontal flipping. We used the Adam optimizer with an initial learning rate of 1.0e-4 and a decay of 1.0e-5 over each update. We set the maximum number of epochs as 100 and the mini-batch size as 16. Besides, the learning rate was reduced by a factor of 10 if the validation loss did not decrease for 5 epochs. The weights were saved from the models with the highest mean AUC on validation set and used for extracting the distinct features of the entire CXR images and local lung region images.

*B.4. Feature fusion module*

Typically, the kernels at the higher layers are learned to encode more abstract features [23]. For each feature extractor in the SDFN, the global average pooling layer is the highest layer and thus contains the most abstract features for thoracic disease classification. To take fully advantage of the distinct features of entire CXR images and local lung region images, the feature fusion module concatenated the global average pooling layers of the two feature extractors. The concatenation layer was further connected to a 14-dimension FC layer with a sigmoid activation function for the final prediction.

Since the two feature extractors in the SDFN have the same input resolution, the image features of the lung regions can be learned at different scales. Consequently, the feature fusion module will learn both the contextual information from the entire CXR images and the multiscale feature representations from the local lung regions.

*B.5. Training strategy*

The proposed SDFN followed a three-stage training strategy. First, we obtained the local lung region images by the LRG. Second, two feature extractors (i.e., fine-tuned DenseNet-121) were separately trained on the entire CXR images and the lung region images. Third, we constructed the SDFN and initialized the feature extractors with their saved weights. During training, only the weights in the feature fusion module were trainable while the saved weights of the feature extractors were fixed. The same hyperparameters and optimization method in section B.3 were used.

*C. Performance assessment*

In this study, two types of deep learning models were involved in terms of their purposes: lung segmentation and disease classification. For the segmentation model, we conducted a five-fold cross-validation on the JSRT Dataset and quantitatively assessed the performance using the Dice similarity coefficient (DSC) and the intersection over union (IoU), as defined by Eq. (3) and (4). Moreover, we qualitatively examined the segmentation results on the Chest X-ray 14 Dataset based on the characteristics described in section B.2.2.

$$DSC = \frac{2|X \cap Y|}{|X| + |Y|} \quad (3)$$

$$IoU = \frac{|X \cap Y|}{|X \cup Y|} \quad (4)$$

where X is the ground truth region and Y is the detected region.

For the disease classification models, we used the benchmark dataset split of the Chest X-ray 14 Dataset for performance assessment. The mean Area under the Receiver Operating Characteristic curve (AUC) of 14 classes was used for the validation purposes. We compared the performance of the proposed SDFN with the fine-tuned DenseNets trained solely on the entire CXR images/local lung region images, as well as other existing methods. The paired t-tests with a p-value 0.01 were conducted to evaluate if the performance improvement of the SDFN was statistically significant.

To assess the lesion detection performance of the SDFN, we followed the same strategy of the class activation map (CAM) [24] to visualize the suspicious lesions, as shown in Figure 5. To generate the CAM for the SDFN, we feed an entire CXR image into the proposed SDFN and extracted the feature maps at the last convolution layers of both feature extractors. Let $f_k(x,y)$ and $f'_k(x,y)$ be the kth feature map at spatial location $(x,y)$ at the last convolution layer of the feature extractor for the entire CXR images and lung region images respectively. Let $\omega_{j,c}$ be the weight between the jth element of the concatenation layer and the cth output in the prediction layer, where $j \in$ [1,2048]. For an activated class c (e.g., the cth pathology), two heatmaps can be obtained by computing the weighted sum of the feature maps using the associated weights, as defined by (5) and (6).

$$heatmap1 = \sum_{j=1}^{1024} \omega_{c,j} f_j(x,y) \quad (5)$$

$$heatmap2 = \sum_{j=1025}^{2048} \omega_{c,j} f'_{j-1024}(x,y) \quad (6)$$

The obtained heatmaps, which originally had the same dimension of the feature maps at the last convolution layer (i.e., 7×7), were resized respectively to the dimensions of the entire CXR and the cropped lung region. Finally, the two resized heatmaps were overlaid according to the cropped location and rescaled to [0,255] for visualization. In our experiment, the detected lesions using the SDFN and the traditional method (i.e., the fine-tuned DenseNet-121 trained solely on the entire CXR images) were generated and compared to the ground truth annotated by a board certified radiologist.

## III. RESULTS

Evaluated on the JSRT Dataset, the five-fold cross-validation showed that the segmentation model achieved a mean DSC of 0.98 and a mean IoU of 0.97. Evaluated on the Chest X-ray 14 Dataset, our experiment showed that the segmentation model successfully identified 95.84% lung regions without false positive/negative segmentation, as shown in Figure 6. It is notable that the identified lung region images largely excluded the noisy regions such as the medical devices and misalignment regions. For the left 4.16% images (4,661 images), we visually examined their cropped lung region images and found most of them successfully enclosed the entire lung regions, as shown in Figure 7. The exceptions were mainly the CXR images with very poor image quality/alignment, and a few non-frontal view CXRs in the dataset (e.g., lateral-view CXRs and leg X-ray images).

In Table 1, we summarized the classification performances of the proposed SDFN and other available methods in terms of AUC scores, evaluated by the NIH benchmark testing set. The ROC curves of the SDFN and the fine-tuned DenseNets trained solely on the entire CXR images or the local lung region images were shown in Figure 8. Our experiment showed that evaluated by the mean AUC over 14 classes, the fine-tuned DenseNet trained on the local lung region images slightly outperformed the one trained on the entire CXR images while the proposed SDFN outperformed both of these fine-tuned DenseNets. Moreover, the SDFN consistently outperformed the traditional method (i.e., the fine-tuned DenseNet trained on the entire CXR images) over all 14 classes. The outperformance was demonstrated to be statistically significant since all the two-tailed p-values in the paired t-tests were below 0.01. It is worth noting that the proposed SDFN achieved a mean AUC score of 0.815 that outperformed all other available approaches [11,14,16].

In addition, our experiment also showed that the proposed SDFN achieved more reliable lesion detection performance compared to the traditional method, as shown in Figure 9. We also demonstrated that not only the pathologies with small lesions (e.g. nodules) could be detected more reliably, but also the ones with relatively larger lesions such as effusion and pneumonia.

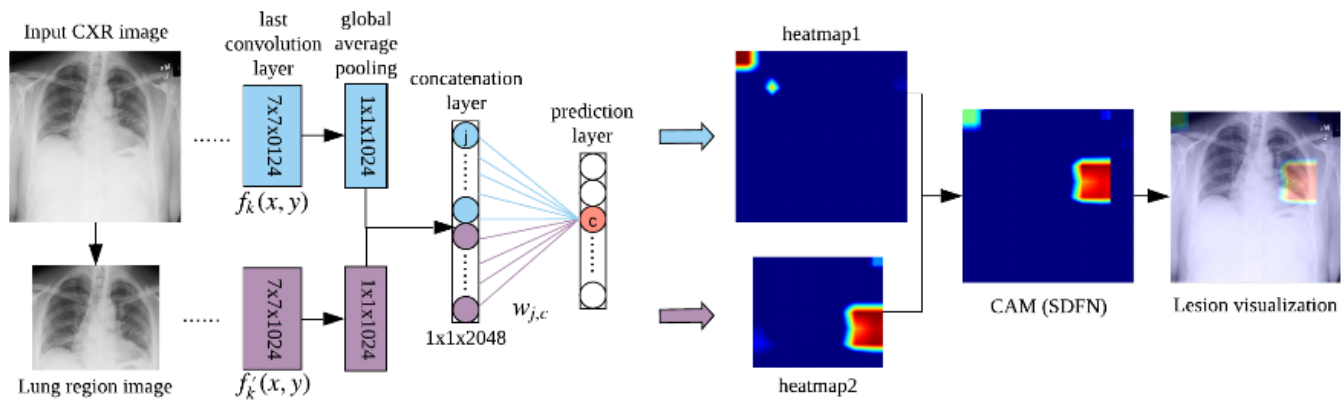

Fig. 5. The flowchart illustrating the generation of the class activation map (CAM) for the proposed SDFN method.

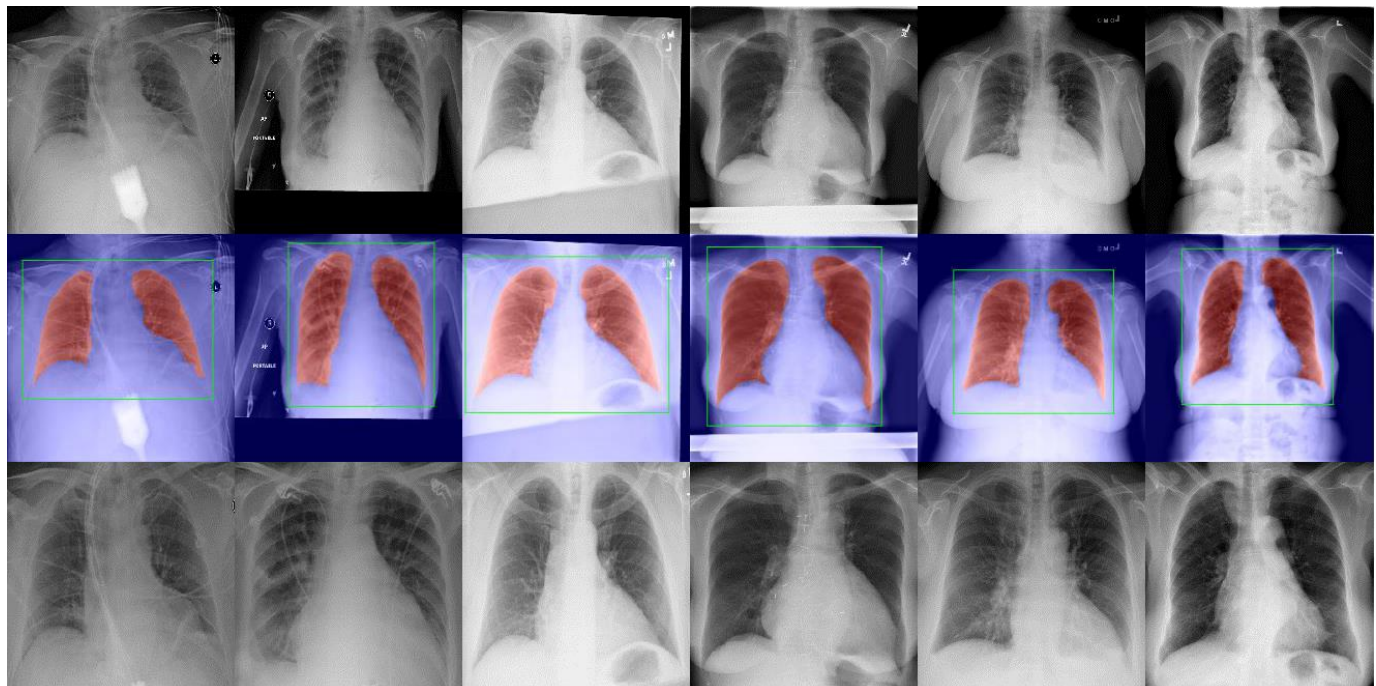

Fig. 6. Examples demonstrating the lung segmentation and the identified lung region images. **Top**: some original CXR images in the Chest X-ray 14 Dataset. **Middle**: the lung segmentation results (in red) and the identified bounding boxes of the lung regions (in green). **Bottom**: the identified lung region images.

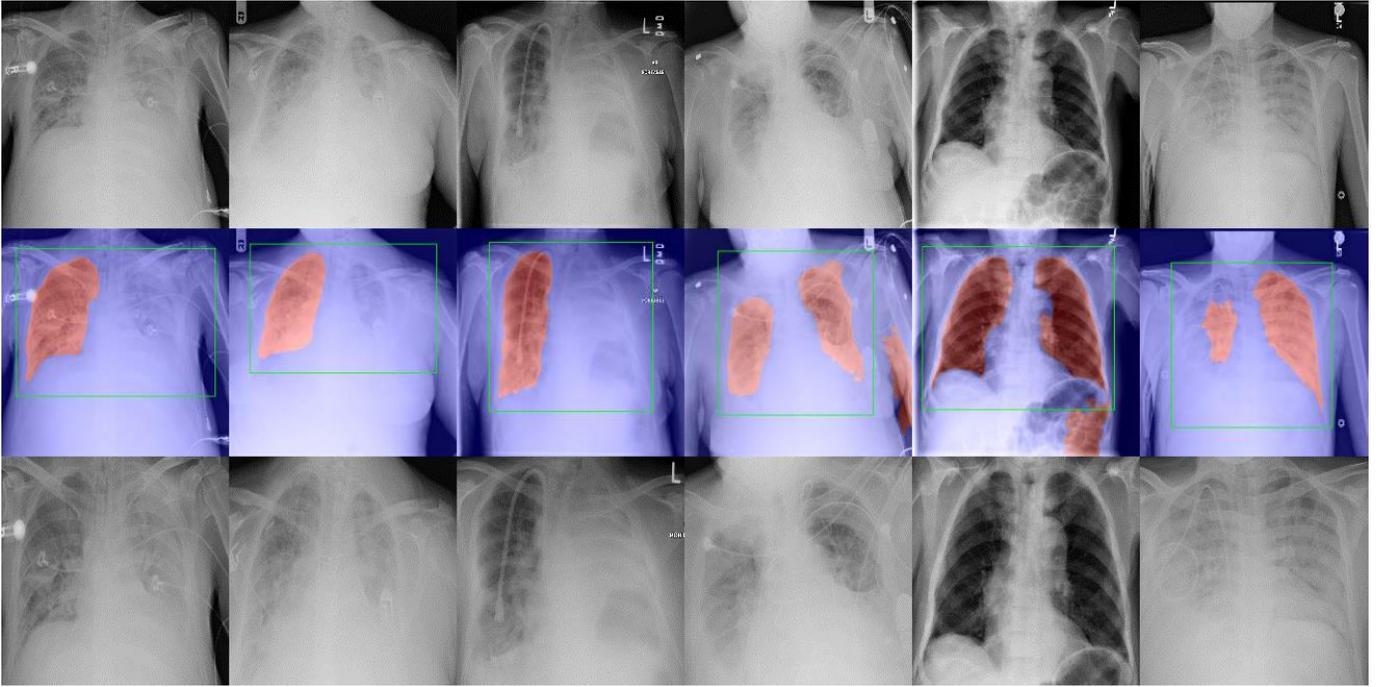

Fig. 7. Examples demonstrating the identified bounding boxes of the lung regions with false positive/negative regions. **Top:** original CXR images. **Middle:** the identified lung segmentation results (in red) and the post-processed bounding boxes (in green). **Bottom:** the identified lung region images.

TABLE I.
PERFORMANCE SUMMARY (AUC SCORES) OF EXISTING APPROACHES AND OUR PROPOSED SDFN USING THE NIH BENCHMARK SPLIT

|  | Wang et al.[11] | Yao et al.[14] | Gündel et al.[16] | Entire CXR | Lung region | Proposed SDFN |
|---|---|---|---|---|---|---|
| Atelectasis | 0.700 | 0.733 | 0.767 | 0.762 | 0.773 | 0.781 |
| Cardiomegaly | 0.814 | 0.856 | 0.883 | 0.878 | 0.868 | 0.885 |
| Effusion | 0.736 | 0.806 | 0.828 | 0.822 | 0.827 | 0.832 |
| Infiltration | 0.613 | 0.673 | 0.709 | 0.693 | 0.670 | 0.700 |
| Mass | 0.693 | 0.777 | 0.821 | 0.791 | 0.807 | 0.815 |
| Nodule | 0.669 | 0.724 | 0.758 | 0.744 | 0.759 | 0.765 |
| Pneumonia | 0.658 | 0.684 | 0.731 | 0.707 | 0.708 | 0.719 |
| Pneumothorax | 0.799 | 0.805 | 0.846 | 0.855 | 0.851 | 0.866 |
| Consolidation | 0.703 | 0.711 | 0.745 | 0.737 | 0.738 | 0.743 |
| Edema | 0.805 | 0.806 | 0.835 | 0.837 | 0.831 | 0.842 |
| Emphysema | 0.833 | 0.842 | 0.895 | 0.912 | 0.909 | 0.921 |
| Fibrosis | 0.786 | 0.743 | 0.818 | 0.826 | 0.825 | 0.835 |
| Pleural Thickening | 0.684 | 0.724 | 0.761 | 0.760 | 0.783 | 0.791 |
| Hernia | 0.872 | 0.775 | 0.896 | 0.902 | 0.912 | 0.911 |
| Mean | 0.740 | 0.761 | 0.807 | 0.802 | 0.804 | **0.815** |

*Entire CXR and Lung region represent the fine-tuned DenseNets trained on the entire CXR images and local lung region images respectively.

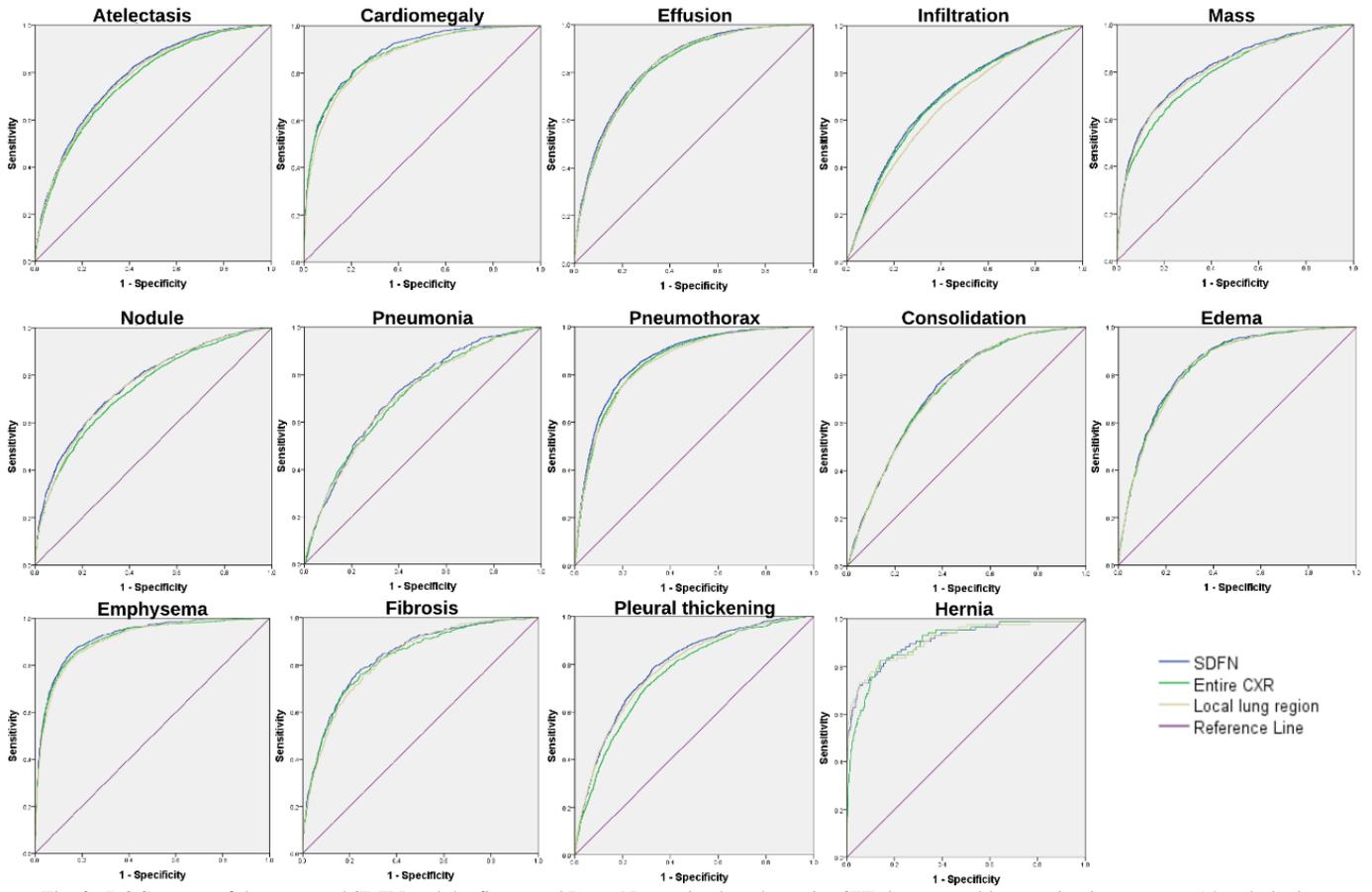
Fig. 8. ROC curves of the proposed SDFN and the fine-tuned DenseNets trained on the entire CXR images and lung region images over 14 pathologies.

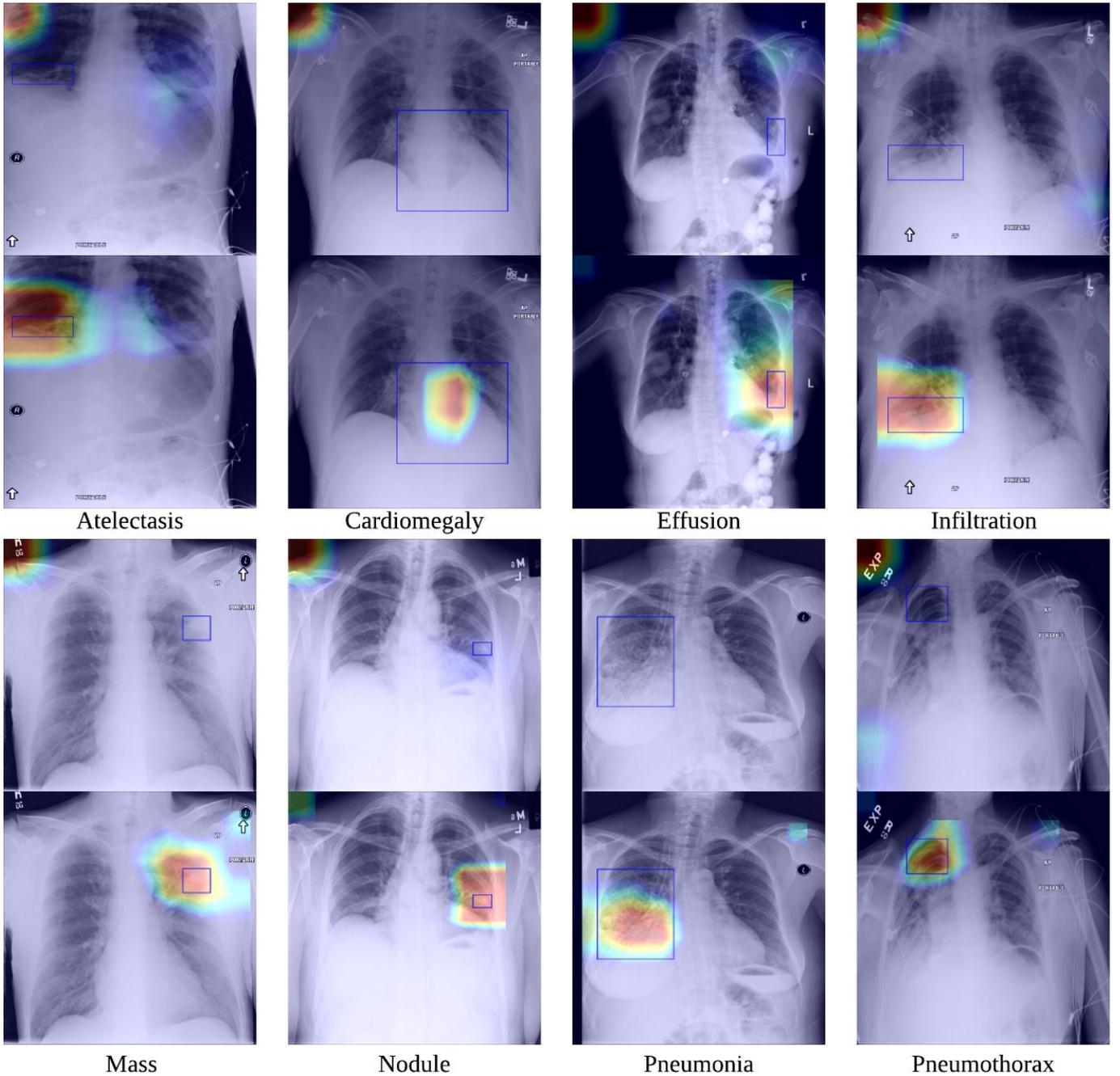

Fig. 9. Examples illustrating the detected lesion regions using the traditional method (i.e., fine-tuned DenseNet-121 trained solely on the entire CXR images) (top) and our proposed SDFN (bottom). The red represents high activation and the blue represents low activation. The blue boxes are the lesion regions annotated by a board certified radiologist.

## IV. DISCUSSION

In this study, we developed a novel deep learning network termed as Segmentation-based Deep Fusion Network (SDFN) to classify 14 thoracic diseases as depicted in CXR images. The most distinctive characteristic of our approach was the fusion of the features from both the entire CXR images and the local lung region images. The novelty of this work lies in the utilization of a multiple scale strategy to improve the classification accuracy. In particular, we demonstrated through this strategy that there might be contextual image features outside the lung regions that suggest the existence of specific diseases.

The primary motivation of this work was the emphasis on the most meaningful part (i.e., the lung regions) of the CXR images to improve the classification accuracy. The lung regions in a CXR image typically occupy around 1/3 or even less of the entire image. If we resize the entire image from 1,024×1,024 pixels to 224×224 pixels for the training purpose, the image details are significantly lost. Hence, we proposed to identify the local lung regions where the pathologies were primarily located. Meanwhile, we also kept the features of the entire CXR images because there might be useful features located outside the lung regions that are associated with specific diseases or suggest their existence. In fact, our experiments suggested this interesting finding since fusing the features of the entire CXR images and the local lung region images achieved a consistently better performance. Moreover, we demonstrated that the proposed SDFN could generate more reliable class activation maps as compared to the traditional method. As illustrated in

Figure 5, the CAMs generated by the proposed method emphasized the activated values of the lung region where the pathologies were primarily located and thus yielded more reasonable and interpretable CAMs. Besides, by leveraging the relatively higher resolution of the lung region images, SDFN could also detect the small lesions that were missed by the traditional methods. In terms of the network structure, the proposed SDFN is similar to the AG-CNN as proposed in [15] since both networks fused the features from the entire images and the local images for final prediction. The difference was that the proposed SDFN used the lung regions as the local images while AG-CNN used the most discriminative region indicated by the CAMs. However, the reliability of these discriminative regions was limited. As shown in Figure 9, a large portion of the most discriminative regions could be located outside the lung regions. Hence, we believe that it is not reliable to simply use the CAM to improve the classification accuracy. In contrast, we proposed to utilize the domain knowledge, namely the thoracic diseases are typically located within the lung regions, to improve the classification accuracy. Compared to the discriminative region indicated by the CAMs, the local lung regions could be reliably identified using the proposed LRG.

In our experiment, the performance of the lung segmentation model on Chest X-ray 14 Dataset was not quantitatively evaluated because (1) there are no available groundtruth lung masks, (2) the objective of the lung segmentation is to identify the bounding box of the lung regions rather than to accurately delineate the lungs. However, our study, which involved more than 100,000 images, showed that the proposed LRG demonstrated a reliable performance in identifying the bounding boxes for lung regions. In this study, we did not quantitatively compare the heatmaps generated by CAMs with the annotated ground-truth, because it is not reasonable to compute the metrics, such as intersection over union (IoU) or mean average precision (mAP), given the fact that the heatmaps or CAMs only have a resolution of 7×7, which is far below the image resolution. Unlike the object detection problems, the heatmaps generated by CAM was used to roughly visualize the most discriminative region instead of to accurately locate / detect the diseases. As we demonstrated in Figure 9, the locations of the CAMs are often incorrect.

One of the interesting findings of our study was that the network trained only on the locally cropped images did not outperform the network on entire CXR images in all the pathologies. However, the integration of the local images and the global images resulted in a consistent higher classification accuracy, suggesting the entire CXR images may contain contextual image features outside of the lung regions that are associated with specific diseases. In addition, it is difficult to always guarantee a proper segmentation of all the lung regions in practice. Due to the existence of a variety of diseases, the lung regions may not be correctly identified from the chest x-ray images. Therefore, it is prudent to combine the local images and the global images to improve both the accuracy and the reliability of the thoracic disease classification.

For the proposed SDFN, we set the input images at a resolution of 224×224 pixels primarily for a fair comparison with available investigations, such as those described in [13, 15]. Typically, training with higher (proper) resolution images yields better performances for deep learning. However, it is notable that training on higher resolution images will necessitate more features involved in the deep learning and a larger dataset. Therefore, in practice, we need to set proper image resolution based on the consideration of a number of factors, such as dataset size and specific applications. For example, in [11, 16], the authors used images with a resolution of 1024×1024 for training purpose, but the performance improvement was very limited.

In our experiment, we found that that the fine-tuned DenseNet trained on the identified lung region images could outperform the one trained on the entire CXR images in certain classes, especially on the pathologies with small lesions such as 'nodule' and 'mass'. The underlying reason may be due to the fact that the local lung region images largely excluded the noisy regions and made effective use of multiscale feature representations. However, some classes such as 'emphysema' were better classified if the network was trained on the entire CXR images, indicating that there were useful features in the entire CXR image but located outside of the lung region. For example, one of the important x-ray findings of emphysema is the hyper inflated lungs, which include the signs such as increased retrosternal air and hyper lucent lung fields [29]. Hence, the intensity contrast between the CXR background (i.e., air) and the lung regions may be a crucial feature for detecting emphysema, but this feature could only be captured by the entire CXR images.

The performance of the proposed SDFN was evaluated by the mean AUC score of 14 classes using the NIH benchmark split on Chest X-ray 14 Dataset. We did not directly compare the SDFN with the methods that randomly split the dataset as performed in [12,13,15], because (1) there may be patient overlap between the training and testing datasets, (2) the classification performances may vary significantly between different splits due to the imbalanced data. Nevertheless, the experimental result showed that the proposed method outperformed all the methods [11,14,16] that were validated on the exact same testing set, as well as the fine-tuned DenseNets trained solely on the entire CXR images or lung region images. Most important, the proposed SDFN achieved higher AUC scores over all class as compared to the traditional method (i.e., fine-tuned DenseNet trained on the entire CXR images). Our paired t-test showed that the outperformance of SDFN was statistically significant, suggesting that the features of the entire CXR images and local lung region images were effectively fused to achieve better performances in both disease classification and lesion detection. We are aware that there might be some unavoidable problems with the dataset labels. However, the number of incorrectly labeled images should be limited as compared to the entire dataset and thus should not affect the development and validation of a computer algorithm, especially when the comparison of different algorithms was performed on the same dataset.

V. CONCLUSION

In this paper, we proposed a novel approach termed as Segmentation-based Deep Fusion Network (SDFN) to classify 14 thoracic diseases. The SDFN initially identified the local lung regions of the entire CXR image by a segmentation model and post-processing. Then the discriminative features of the entire CXR images and the local lung region images were extracted separately and fused for the final prediction. Using the NIH benchmark split on Chest X-ray 14 Dataset, our experiment demonstrated that the proposed method achieved more accurate disease classification performance as compared with the available methods and more reliable lesion detection performance compared to the traditional method.

ACKNOWLEDGEMENTS

This work is supported by National Institutes of Health (NIH) (Grant No. R21CA197493, R01HL096613) and Jiangsu Natural Science Foundation (Grant No. BK20170391).

REFERENCES

[1]. Brady A, Laoide RÓ, McCarthy P, McDermott R, "Discrepancy and Error in Radiology: Concepts, Causes and Consequences", *the Ulster Medical Journal*. 2012; 81(1):3-9.
[2]. H. Fu, J. Cheng, Y. Xu, C. Zhang, D. Wong, J. Liu, X. Cao, "Disc-aware Ensemble Network for Glaucoma Screening from Fundus Image," in IEEE Transactions on Medical Imaging.
[3]. H. Pratt, F. Coenen, D. M. Broadbent, S. P. Harding, Y. Zheng, "Convolutional neural networks for diabetic retinopathy", *Procedia Computer Science*, vol. 90, pp. 200-205, Jul. 2016.
[4]. M. Anthimopoulos, S. Christodoulidis, L. Ebner, A. Christe and S. Mougiakakou, "Lung Pattern Classification for Interstitial Lung Diseases Using a Deep Convolutional Neural Network," in *IEEE Transactions on Medical Imaging*, vol. 35, no. 5, pp. 1207-1216, May 2016.
[5]. P. Nardelli, D. Jimenez-Carretero, D. Bermejo-Pelaez, G. R. Washko, F. N. Rahaghi, M. J. Ledesma-Carbayo, and R. Estépar, "Pulmonary Artery-Vein Classification in CT Images Using Deep Learning", in IEEE Transactions on Medical Imaging. 2018.
[6]. F. Commandeur, M. Goeller, J. Betancur, S. Cadet, M. Doris, X. Chen, D. S. Berman, P. J. Slomka, B. K. Tamarappoo, and D. Dey, "Deep Learning for Quantification of Epicardial and Thoracic Adipose Tissue from Non-Contrast CT", in IEEE Transactions on Medical Imaging. 2018.
[7]. E. Nasr-Esfahani, S. Samavi, N. Karimi, S.M.R. Soroushmehr, K. Ward, M.H. Jafari, B. Felfeliyan, B. Nallamothu, and K. Najarian, "Vessel extraction in X-ray angiograms using deep learning," 2016 38th Annual International Conference of the IEEE Engineering in Medicine and Biology *Society (EMBC)*, Orlando, FL, 2016, pp. 643-646.
[8]. P. Liskowski and K. Krawiec, "Segmenting Retinal Blood Vessels With Deep Neural Networks", in *IEEE Transactions on Medical Imaging*, vol. 35, no. 11, pp. 2369-2380, Nov. 2016.
[9]. X. Yang, R. Kwitt, M. Styner, and M. Niethammer, ''Quicksilver: Fast predictive image registration—A deep learning approach,'' Neuroimage, vol. 158, pp. 378–396, 2017
[10]. S. Miao, Z. J. Wang, and R. Liao, ''A CNN regression approach for real-time 2D/3D registration,'' IEEE Trans. Med. Imag., vol. 35, no. 5, pp. 1352–1363, 2016.
[11]. X. Wang, Y. Peng, L. Lu, Z. Lu, M. Bagheri, R. Summers, "ChestX-ray8: Hospital-scale Chest X-ray Database and Benchmarks on Weakly-Supervised Classification and Localization of Common Thorax Diseases", *IEEE CVPR*, pp. 3462-3471, 2017.
[12]. L. Yao, E. Poblenz, D. Dagunts, B. Covington, D. Bernard, and K. Lyman, "Learning to diagnose from scratch by exploiting dependencies among labels", *arXiv preprint arXiv*: 1710.10501, 2017.
[13]. P. Rajpurkar, J. Irvin, K. Zhu, B. Yang, H. Mehta, T. Duan, D. Ding, A. Bagul, C. Langlotz, K. Shpanskaya, M. P. Lungren, A. Y. Ng, "CheXnet: Radiologist-level pneumonia detection on chest x-rays with deep learning", *arXiv preprint arXiv*: 1711.08760, 2017.
[14]. L. Yao, J. Prosky, E. Poblenz, B. Covington and K. Lyman, "Weakly supervised medical diagnosis and localization from multiple resolutions", arXiv preprint arXiv: 1803.07703. 2018.
[15]. Q. Guan, Y. Huang, Z. Zhong, Z. Zheng, L. Zheng, Y. Yang, "Diagnose like a radiologist: attention guided convolutional neural network for thorax disease classification", *arXiv preprint arXiv*: 1801.09927, 2018.
[16]. S. Gündel, S. Grbic, B. Georgescu, K. Zhou, L. Ritschl, A. Meier, and D. Comaniciu, "Learning to recognize abnormalities in chest x-rays with location-aware dense networks", *arXiv preprint arXiv*: 1803.04565. 2018.
[17]. G. Huang, Z. Liu, L. v. d. Maaten and K. Q. Weinberger, "Densely Connected Convolutional Networks," *2017 IEEE Conference on Computer Vision and Pattern Recognition (CVPR)*, Honolulu, HI, 2017, pp. 2261-2269.
[18]. J. Shiraishi, S. Katsuragawa, J. Ikezoe, T. Matsumoto, T. Kobayashi, K. Komatsu, M. Matsui, H. Fujita, Y. Kodera, and K. Doi, "Development of a digital image database for chest radiographs with and without a lung nodule: receiver operating characteristic analysis of radiologists' detection of pulmonary nodules", American Journal of Roentgenology, vol. 174, p. 71-74, 2000.
[19]. B. van Ginneken, M.B. Stegmann, M. Loog, "Segmentation of anatomical structures in chest radiographs using supervised methods: a comparative study on a public database", Medical Image Analysis, nr. 1, vol. 10, pp. 19-40, 2006.
[20]. O. Ronneberger, P. Fischer, T. Brox, "U-Net: Convolutional Networks for Biomedical Image Segmentation", In: Navab N., Hornegger J., Wells W., Frangi A. (eds) Medical Image Computing and Computer-Assisted Intervention – MICCAI 2015. MICCAI 2015. Lecture Notes in Computer Science, vol 9351. Springer, Cham.
[21]. Pazhitnykh I., Petsiuk V.: Lung Segmentation (2D), https://github.com/imlabuiip/lung-segmentation-2d
[22]. J. Deng, W. Dong, R. Socher, L. J. Li, Kai Li and Li Fei-Fei, "ImageNet: A large-scale hierarchical image database", *2009 IEEE Conference on Computer Vision and Pattern Recognition*, Miami, FL, 2009, pp. 248-255.
[23]. J. Gu, Z. Wang, J Kuen, L. Ma, A. Shahroudy, B. Shuai, T. Liu, X. Wang, L. Wang, G. Wang, J. Cai and T. Chen, "Recent Advances in Convolutional Neural Networks," *arXiv preprint arXiv*: 1512.07108. 2015.
[24]. B. Zhou, A. Khosla, A. Lapedriza, A. Oliva and A. Torralba, "Learning Deep Features for Discriminative Localization," *2016 IEEE Conference on Computer Vision and Pattern Recognition (CVPR)*, Las Vegas, NV, 2016, pp. 2921-2929.
[25]. A. Krizhevsky, L. Sutskever, G. E. Hinton, "Imagenet classification with deep convolutional neural networks", in Advances in neural information processing systems, 2012, pp. 1097-1105.
[26]. C. Szegedy, W. Liu, Y. Jia, P. Sermanet, S. Reed, D. Anguelov, D. Erhan, V. Vanhoucke, and A. Rabinovich, "Going deeper with convolutions", *2015 IEEE Conference on Computer Vision and Pattern Recognition (CVPR)*, Boston, MA, 2015, pp. 1-9.
[27]. K. Simonyan and A. Zisserman, "Very deep convolutional networks for large-scale image recognition", *arXiv preprint arXiv*: *1409.1556*, 2014.
[28]. K. He, X. Zhang, S. Ren, and J. Sun, "Deep residual learning for image recognition", in *Proceedings of the IEEE conference on computer vision and pattern recognition*, 2016, pp. 770-778.
[29]. A. Chandrasekharhttp.: chronic obstructive lung disease (COPD)/emphysema: www.stritch.luc.edu/lumen/MedEd/Radio/curriculum/Medicine/emphysema.htm (2018).